\documentclass[lettersize,journal]{IEEEtran}
\usepackage{amsmath,amsfonts}
\usepackage{algorithmic}
\usepackage{algorithm}
\usepackage{array}
\usepackage[caption=false,font=normalsize,labelfont=sf,textfont=sf]{subfig}
\usepackage{textcomp}
\usepackage{stfloats}
\usepackage{url}
\usepackage{verbatim}
\usepackage{graphicx}
\usepackage{cite}
\usepackage[normalem]{ulem}
\usepackage{orcidlink}
\usepackage{url}
\begin{document}

\title{HydroVision: Predicting Optically Active Parameters in Surface Water Using Computer Vision}

\author{%
\begin{tabular}{c}
Shubham Laxmikant Deshmukh\orcidlink{0009-0005-3399-0703}\textsuperscript{1},
Matthew Wilchek\orcidlink{0000-0001-8664-1586}\textsuperscript{1},
Feras A. Batarseh\orcidlink{0000-0002-6062-2747}\textsuperscript{2} \\[0.4ex]
\textsuperscript{1}Department of Computer Science, Virginia Tech, Arlington, VA, USA \\
\textsuperscript{2}Department of Biological Systems Engineering, Virginia Tech, Arlington, VA, USA \\[0.3ex]
\texttt{shubhamd23@vt.edu, mwilchek@vt.edu, batarseh@vt.edu}
\end{tabular}
}

\maketitle

\begin{abstract}
Ongoing advancements in Computer Vision, particularly in pattern recognition and scene classification, have paved the way for innovative environmental monitoring applications. Deep Learning has demonstrated promising results, enabling non-contact water quality monitoring and contamination assessment—both essential for disaster response and public health protection. This manuscript proposes HydroVision, a novel deep learning-based scene classification framework that estimates optically active water quality parameters, such as Chlorophyll-$\alpha$, Chlorophylls, Colored Dissolved Organic Matter (CDOM), Phycocyanins, Suspended Sediments, and Turbidity, from Red-Green-Blue (RGB) images of surface water. The utility of this model lies in early detection of contamination trends and supporting monitoring efforts by regulatory bodies during external environmental factors, industrial activities, and force majeure events. It is trained on a diverse and extensive dataset of over 500,000 seasonally varied images sourced from the United States Geological Survey (USGS) Hydrologic Imagery Visualization and Information System (HIVIS) database between early 2022 and late 2024. The proposed model introduces an innovative approach to water quality monitoring and assessment using widely available RGB imagery, serving as a scalable and cost-effective alternative to traditional multispectral/hyperspectral remote sensing. The model is trained using four state-of-the-art Convolutional Neural Network (CNN) architectures—VGG-16, ResNet50, MobileNetV2, DenseNet121—and a Vision Transformer (ViT), leveraging transfer learning to determine the optimal framework for predicting six optically active parameters. Among them, the best-performing model achieves a validation $R^2$ (coefficient of determination) score of 0.89 using DenseNet121 for predicting CDOM, underscoring its potential for real-world water quality assessment across diverse environmental conditions. Although trained on well-lit images, enhancing robustness under low-light or obstructed conditions offers a promising direction for expanding its practical utility.
\end{abstract}

\begin{IEEEkeywords}
Water Quality, Computer Vision, Deep Learning, Surface Water, Optical Parameters, Remote Sensing.
\end{IEEEkeywords}

\section{Introduction}
\label{sec:Intro}
Covering approximately 71\%  of the planet's surface, water is found in oceans, rivers, lakes, ponds, and even as atmospheric moisture. However, of this vast coverage, a staggering 97.5\% comprises ocean and other saline water, leaving only 2.5\% as freshwater available for human and ecological use \cite{usgs_water_distribution}. The fraction of water truly accessible for human consumption is alarmingly small \cite{epa_water_use}. The global water crisis is worsening due to population growth. By 2030, the demand for water could exceed the supply by 40\% \cite{worldbank_water_management}. Scarcity, extreme weather, and hydrological unpredictability threaten economic stability and can even spark conflicts \cite{divaenvitec_water_conflict}. 

With such limited freshwater resources available, it is imperative to develop effective water management strategies. A critical component of such strategies are robust water quality monitoring systems, which assess the condition of surface water bodies such as rivers, lakes, and ponds. These systems enable proactive water safety assessments and swift intervention measures, ensuring the sustainable and secure use of water resources—especially during critical emergencies such as accidental contamination or force majeure events like flooding. Monitoring surface water quality is not only a technical endeavor, but a necessity for safeguarding the health of ecosystems and communities alike \cite{CHAPMAN2022132}.

Water quality can be assessed through a wide range of physical, chemical, and biological parameters\cite{epa_water_quality}. However, each parameter is of separate importance, and different collection methods are used with different units. A combination of these multiple parameters can provide a definite answer as to whether a given water body is safe for human use or not. Out of these parameters, there is a different subset, namely: optically active parameters \cite{Ahmed2022}. Optically active water quality parameters measure how substances in a water sample interact with light, particularly by scattering it when exposed to a beam of light \cite{gholizadeh2016comprehensive}. This interaction between the concentration of optically active substances in the water and the color of the water affects water quality. The most common optically active parameters used for water quality monitoring are turbidity, total suspended solids, and Chlorophyll-$\alpha$~\cite{Ahmed2022}. Optically active indicators are particularly valuable for non-contact, image-based assessments, as they respond to how light interacts with the water's surface. Their optical characteristics such as reflectance, color, and turbidity can be effectively analyzed using computer vision techniques and convolutional neural network (CNN) models. This study explores the use of such models to predict optically active parameters directly from Red-Green-Blue (RGB) images.

This paper is structured as follows: Section~\ref{sec:Motivation} discusses the motivation behind using optically active parameters for water quality analysis. Section~\ref{sec:Research Questions and Contributions} outlines the key research questions addressed in this study and highlights the contributions of this manuscript to the research community.
Section~\ref{sec:RelatedWork} reviews related work, providing an overview of existing applications of Computer Vision in water quality analysis. Section~\ref{sec:Data Collection and Management} describes a carefully curated dataset designed for future research on evaluating scene classification models with water images, along with the steps taken for its collection and segmentation. Section~\ref{sec: Methods} outlines the methods used in our study, including model design and selection of Convolutional Neural Network (CNN) models. Section~\ref{sec:Results} presents the results and evaluation metrics of HydroVision. Finally, Section~\ref{sec:Discussion} offers concluding remarks, limitations and discusses potential future work.

\subsection{Motivation}
\label{sec:Motivation}
Water quality is generally defined by its suitability for specific uses, which depends on selected physical, chemical, and biological characteristics \cite{usgs_water_quality_use}. These parameters are assessed to indicate ecosystem health, safety for human use, and pollution levels. Under the Clean Water Act (CWA) \cite{cornell_clean_water_act}, various water quality parameters are recommended by state governments. The United States Environmental Protection Agency (USEPA) highlights some of the most common parameters, including temperature, dissolved oxygen, pH, turbidity, macroinvertebrates, habitat assessment, and other chemical/biological indicators \cite{usgs_water_quality_2023,epa_water_quality_standards_2023}. However, traditional water quality monitoring systems face challenges such as high operational costs, significant maintenance demands, and scalability concerns \cite{watertech_monitoring_challenges_2023}. These limitations underscore the need for cost-effective, non-contact, and resilient methods for predicting water quality parameters, especially during emergencies. In particular, image-based monitoring of optically active parameters offers a practical, scalable approach for early detection using Computer Vision models. RGB-based Computer Vision approaches meet this need by utilizing widely available sensors, enabling practical and large-scale deployment without the cost or complexity of hyperspectral systems \cite{zhu2022mlwater}.

Optically active parameters—such as chlorophyll-$\alpha$, chlorophylls, phycocyanins, suspended sediments, and turbidity—use light-based methods to measure wavelengths absorbed or emitted by constituents in water. For example, turbidity is typically measured by the scattering of light, producing values in NTU (Nephelometric Turbidity Units), an indirect indicator of pollution. However, factors such as temperature, particle size, and chemical interferences can affect readings, requiring calibration and correction procedures \cite{C5EM00030K}.

Turbidity serves as a critical indicator of water quality and safety, signaling potential contamination and the need for preventive action \cite{usgs_turbidity}. This makes RGB-based AI models particularly valuable for timely water quality assessments in scenarios where rapid action is critical. For instance, on July 3, 2024, D.C. Water and the Army Corps of Engineers issued a water advisory for Washington, D.C., due to increased turbidity affecting production at the Dalecarlia Water Treatment Plant \cite{arlnow_boiling_advisory,dcwater_boil_advisory}. Similarly, on September 30, 2024, the New River Valley Regional Water Authority issued a boil water notice in response to severe flooding that raised turbidity beyond regulatory limits in Blacksburg, Virginia, and Montgomery County, Maryland \cite{cardinalnews_boiling_advisory}.

External events such as urban expansion, industrial discharge, and accidents can rapidly deteriorate surface water quality. On January 30, 2025, a collision between a commercial plane and a military helicopter over the Potomac River posed a potential contamination risk near Washington, D.C.'s water infrastructure, due to potential fuel and debris leakage into the surface water system. \cite{nytimes2025planecrash}.

Natural disasters (e.g., flooding, earthquakes) also pose substantial risks to water security \cite{who2025drinkingwater}. Ensuring access to safe drinking water during such events is critical \cite{aquasana2025disasters}. Rapid assessment methods driven by artificial intelligence (AI) are essential for emergency response and public health protection \cite{usgs2025urbanization}. While RGB imagery lacks the spectral depth of hyperspectral sensors, it is far more accessible, cost-effective, and scalable—making it practical for near-real-time monitoring, especially when enhanced with AI models trained on diverse environmental datasets. However, RGB-based monitoring can be sensitive to environmental lighting conditions such as cloud cover, shadows, or low-light scenarios, which may introduce variance in model predictions. Addressing such challenges is key to ensuring model robustness across diverse operational settings. Recent advances in deep learning have enabled accurate estimation of optically active parameters from RGB imagery, narrowing the performance gap compared to multispectral data in relevant tasks \cite{zhu2017deeplearning}.

Recent literature supports AI applications in real-time water quality monitoring \cite{Sreng2024WaterSystemsSecurity}, cyberbiosecurity \cite{sobien2023aiforcybersecurity}, AI-based forecasting \cite{kulkarni2023p2o}, generative anomaly detection \cite{sikder2023deepH2O}, flood prediction \cite{lin2023neuralflood}, and integrated AI testbeds for sustainable water and agriculture management \cite{DrB2023acwa}.

\subsection{Research Questions and Contributions}
\label{sec:Research Questions and Contributions}
Safeguarding water resources requires reliable, cost-effective, and scalable monitoring tools that remain functional under external stressors like industrial accidents, environmental contamination, and natural disasters. The need for resilient, non-contact monitoring solutions is pressing, especially in emergency scenarios where rapid response is essential.

AI-powered Computer Vision offers significant promise in this context. In particular, it enables estimation of optically active water parameters using visual cues in RGB images—without relying on expensive or inaccessible remote sensing tools. By analyzing RGB images of surface water, convolutional neural networks (CNNs) and other Vision Transformer (ViT) models can identify visual patterns—such as cloudiness, sediment load, and surface anomalies—associated with optically active parameters. These models enable real-time, low-cost, and remote water quality assessment using widely accessible RGB imagery, though with certain limitations in detecting parameters beyond visual cues.

To address current limitations and guide this investigation, we pose the following research questions:
\begin{enumerate}
    \item \textbf{RQ1:} What are the challenges in sourcing and preparing high-quality, accurately labeled datasets for training environmental monitoring scene classification models?
    \item \textbf{RQ2:} How effectively can AI-based Computer Vision techniques identify and quantify optically active water quality parameters from RGB images under varying environmental conditions?
\end{enumerate}

These insights have critical real-world applications. Rapid contamination assessments following industrial spills, aircraft incidents, or flooding events can enable informed decision-making and ensure access to safe water. Timely detection of anomalies supports both environmental safety and public health.

Building on this premise, our contributions are twofold:
\begin{itemize}
    \item We introduce a processed and cleaned benchmark dataset derived from the United States Geological Survey (USGS) Hydrologic Imagery Visualization and Information System (HIVIS) database, capturing seasonal variation across 111 monitoring sites between early 2022 and late 2024.
    \item We develop five CNN-based models and one ViT model trained on RGB imagery to predict six optically active parameters, achieving strong validation performance ($\mathbf{R^2} = 0.898$, coefficient of determination) using DenseNet121 for Colored Dissolved Organic Matter (CDOM).
\end{itemize}

\section{Related Work}
\label{sec:RelatedWork}
Numerous studies have leveraged Machine Learning (ML) and Deep Learning techniques to analyze, monitor, and predict water quality parameters \cite{essamlali2024waterquality}. However, the application of Computer Vision techniques to water quality assessment has predominantly occurred within controlled settings rather than in natural, real-world environments \cite{liu2018review}. This section reviews the existing literature on the application of computer vision methodologies to monitor water quality through optically active parameters. We begin by discussing the scope and outcomes of these methods in controlled studies, then by examining significant research that has ventured into applying these techniques in practical, real-world applications.

In our review, we examined approximately 22 papers, which highlighted the most commonly studied optically active variables and techniques for assessing water quality, including:
 i) turbidity, ii) total suspended solids, iii) image processing techniques, and iv) biological sensors. Using Computer Vision and CNN architectures, these parameters are utilized to predict, analyze, or monitor the water quality of water bodies, both in controlled environments and open settings.

\subsection{AI Methodologies for Water Quality Analysis}
Turbidity is an important indicator of water quality, reflecting how cloudy or hazy water appears due to suspended particles. This optical property is measured by observing how light is scattered when it passes through the water. Higher turbidity occurs when more light is scattered, which is caused by a greater concentration of suspended particles. Thus, an increase in turbidity typically signifies a rise in water pollution \cite{feizi2023image}.

Given its optical basis, significant research has focused on predicting turbidity using the RGB values of water images. Most approaches begin by pre-processing the images through Computer Vision techniques such as data augmentation, image denoising, and resizing, then dividing the dataset into training and testing sets. After pre-processing, the images are input into convolutional neural networks (CNNs) for training, allowing the models to learn relationships between spatial pixel patterns and turbidity measurements \cite{liu2018review}.

The CNN architectures employed range from simpler, custom models with a few convolutional layers to complex, widely recognized networks like ResNet and VGG-16, often used in transfer learning for enhanced pattern recognition. Most datasets in these studies are relatively small, typically containing a few hundred images captured in controlled environments. The turbidity levels are often classified into categories such as Excellent, Acceptable, Slightly Polluted, Polluted, and Highly Polluted, each associated with specific NTU ranges or a binary classification of contaminated or non-contaminated water \cite{youssef2023estimation,montassar2020water}.

Some studies have used larger data sets that contain thousands of images, using controlled environment techniques indoors, such as adding clay to simulate varying levels of turbidity. These studies often focus on a specific region of interest in the images to enhance accuracy in predicting turbidity ranges \cite{jantarakasem2024estimating,zhou2024approach,liu2018review}. Additionally, data collection methods vary widely, including satellite imagery, aerial perspectives (e.g., drones), and eye-level perspectives from handheld or stationary cameras. These different viewpoints introduce variations in context, background clutter, and lighting conditions, which can influence model performance and highlight the need for robust training datasets that account for diverse scenarios.

Another optically active parameter for water quality assessment is Total Suspended Solids (TSS). TSS measures the concentration of suspended particles in a given volume of water, which can include substances such as sand, sediments, and plankton—typically particles larger than 2 microns that do not dissolve in water \cite{totalSuspendedSolids2024}. Due to its optical properties, TSS can be an indicator of water contamination, with higher TSS (measured in mg/L) suggesting a greater level of pollution or safety risk in the water. 

Computer Vision techniques, such as Otsu thresholding, morphological operations, and edge detection, can effectively detect sediment particles before passing the images to a Deep Learning model. In studies, image datasets or video sequences are often captured in controlled environments using smartphones and then fed into advanced Deep Learning architectures like custom CNNs, VGG-16, VGG-19, Inception-V3, AlexNet, Xception, DenseNet-121, MobileNet, and NASNet-Mobile to classify different levels of suspended solids \cite{Hacıefendioğlu2023,Lopez-Betancur2022, ShengZhang2019}.

\subsection{Real-World Applications of CV in Water Quality Analysis}
Research predicting optically active water quality measures in real-world environments is limited. To date, only one study has used open-environment water surface images to detect turbidity values through the open-source USGS HIVIS platform \cite{zhou2024approach}. In this study, researchers pre-classified images based on favorable lighting conditions before applying transfer learning with the ResNet50 architecture to predict turbidity in Formazin Nephelometric Units (FNU). The same study also proposed a video image-based turbidity recognition model for continuous, non-contact monitoring using deep learning and real-time image analysis.

Similarly, few studies have focused on predicting Total Suspended Solids (TSS) in real-world water body images. One notable study utilized the EfficientNet architecture to classify TSS levels, predicting whether values were above or below 5 mg/L \cite{AntunezDurango2023}.

In addition to core water quality parameters, researchers have investigated impurity detection in surface water, such as algae detection. These studies heavily rely on image pre-processing techniques in Computer Vision, including color space conversion from Blue-Green-Red (BGR) to Hue-Saturation-Value (HSV), segmentation, contour detection, and feature extraction methods such as Gabor filters and Local Binary Patterns (LBP). ML models are trained to classify or detect impurities using these extracted features and color information \cite{10.1016/j.engappai.2023.106376}. The models applied in these studies range from traditional approaches such as XGBoost, Linear Regression, Support Vector Regression (SVR), Ridge Regression, K-Nearest Neighbors (KNN), and K-Means clustering to advanced CNN models like Region-based Fully Convolutional Network (R-CNN), Faster R-CNN, ShuffleNetV2, and Single Shot MultiBox Detector (SSD) \cite{samantaray2018algae,sharma2021water,erfani2023vision,gupta2019aquasight,ruiz-navarro2022computer,zhang2024machine}.

Researchers have also explored innovative approaches to predict water quality using biological sensors, such as aquatic organisms or fish. The specific movement patterns or rhythmic behaviors of these organisms can serve as natural bioindicators of water quality. By observing and analyzing these movements with Computer Vision techniques, researchers can utilize time-series Deep Learning models, such as Long Short-Term Memory (LSTM) and Recurrent Neural Networks (RNN), to predict water quality based on these natural indicators \cite{hu2022computer, zheng2014method, yuan2018biological, horak2015water}.

\section{Data Collection and Management}
\label{sec:Data Collection and Management}
This section outlines the processes involved in curating our dataset for scene classification with a focus on water body images. It covers data acquisition through the latest United States Geological Survey (USGS) resources available to the public, image segmentation techniques, and filtering methods used to enhance dataset quality. Additionally, it introduces a data dictionary designed to facilitate efficient curation of train/test/validation splits, with the flexibility to map metadata to images for future analysis.

\subsection{USGS Data Collection}
\label{subsec:DataCollect}
USGS is known for making geological data on various U.S. locations publicly accessible for research. In February 2022, USGS introduced the Hydrologic Imagery Visualization and Information System (HIVIS), a new platform providing access to images from over 840 water resource monitoring sites nationwide. Each site has a unique image capture frequency; for example, the Platte River monitoring site near Grand Island, Nebraska, captures images and water parameters every 15 minutes, while the Green Bay Oil Depot site in Wisconsin collects data every five minutes \cite{usgs_hivis}.

USGS also provides a public portal for water data called the National Water Information System (NWIS) \cite{usgs_nwis}. The NWIS web interface offers real-time access to data from over 13,500 water stations distributed across all 50 states in the U.S. NWIS encompasses data on surface water, groundwater, water quality, and water use. In this research, we focus specifically on water stations monitoring optically active parameters to assess surface water quality.

The data collection process includes two main components: i) image collection via USGS HIVIS Amazon Web Services (AWS) Application Programming Interface (API), and ii) optically active parameter data collection through USGS's NWIS API. Before data collection, a list of optically active parameters from those available at water monitoring stations is curated. The final list of 6 parameters includes: ``Chlorophyll-$\alpha$'', ``Chlorophylls'', ``Colored Dissolved Oxygen Matter (CDOM)'', ``Phycocyanin'', ``Suspended Sediments'', and ``Turbidity'', with varying units of the same parameter across sites due to differences in measurement tools making a total of 16 unique optically active parameters \cite{usgs_parameters}. Only sites with at least one of these parameters are selected, resulting in 111 water monitoring locations with the relevant data.
%US map figure%

\begin{figure}[h]
    \centering
    \includegraphics[width=\linewidth]{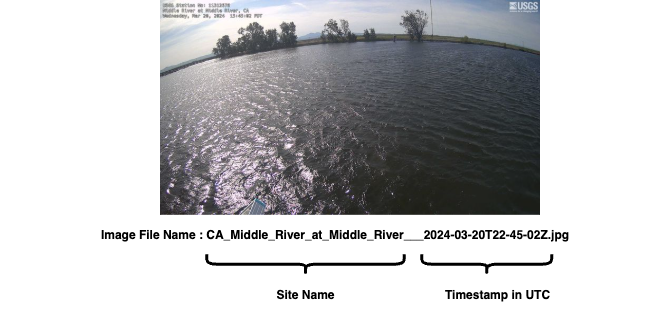} 
    \caption{ An example of Image File naming format collected through USGS HIVIS AWS API.}
    \label{fig:image_label}
\end{figure}

For image collection, the HIVIS API is used to retrieve all available images from 111 sites, starting from the initial camera installation date at each water station (February 2022) until November 2024. The API required site location names, which are obtained by mapping the site codes to their respective HIVIS portal-specific names. A large download limit is configured to capture the complete image history, which is stored as `.jpg` files on Amazon AWS \cite{usgs_hivis_apps}. The captured image file names followed the format: \texttt{Site\_Name\_\_Timestamp.jpg}, where the timestamps are recorded in Coordinated Universal Time (UTC) format as shown in Fig.~\ref{fig:image_label}.

Parameter data are collected using USGS's NWIS Water Data API \cite{usgs_nwis}, which provides access to real-time and historical data from monitoring sites across the U.S. The API is configured with filtered site codes and parameter codes relevant to this research, along with a specified date range from January 1, 2018, to November 14, 2024. The retrieved dataset, which includes features and target variables, is saved in CSV format, containing columns for the site code, site name, timestamp, and values for 16 optically active parameters. A detailed description of these columns is provided in Table~\ref{tab:csv_columns}, which outlines the structure of the dataset and the measurement units used for each parameter. These parameters are critical for analyzing water quality across various monitoring sites.

\begin{table}[h]
    \centering
    \caption{Description of Features and Target Variables in the Dataset}
    \label{tab:csv_columns}
    \begin{tabular}{|l|p{0.6\linewidth}|} % Adjusted column width
        \hline
        \textbf{Column Name} & \textbf{Description} \\
        \hline
        Site Code & A unique identifier assigned to each monitoring site. \\
        \hline
        Site Name & The full name of the monitoring site associated with the site code. \\
        \hline
        Timestamp & The date and time of the recorded parameter values, formatted in Coordinated Universal Time (UTC) with a time zone offset. \\
        \hline
        Chlorophyll-$\alpha$ & Measured in two units: micrograms per liter ($\mu$g/L) and relative fluorescence units (RFU). \\
        \hline
        Chlorophylls & Measured in three units: micrograms per liter ($\mu$g/L) at 650-700 nm, micrograms per liter ($\mu$g/L) at 470-685 nm, and relative fluorescence units (RFU). \\
        \hline
        CDOM & Measured in two units: parts per billion quinine sulfate equivalents (ppb QSE) and relative fluorescence units (RFU). \\
        \hline
        Phycocyanin & Measured in two units: micrograms per liter ($\mu$g/L) and relative fluorescence units (RFU). \\
        \hline
        Suspended Sediments & Measured in three units: tons per day, milligrams per liter (mg/L) as estimated by regression, and milligrams per liter (mg/L) at a fixed point in the stream. \\
        \hline
        Turbidity & Measured in four units: formazin nephelometric units (FNU), formazin backscatter units (FBU), styrene divinylbenzene backscatter units (SBU), and nephelometric turbidity units (NTU). \\
        \hline
    \end{tabular}
\end{table}

\subsection{Image Segmentation}
The images are collected through USGS's HIVIS AWS API and often include extraneous features such as dams, boats, and grasslands, which are irrelevant to the analysis of surface water bodies. We employ a pre-trained surface water body segmentation model to isolate water regions based on the U-shaped Convolutional Neural Network (U-Net) architecture \cite{kaggle_segmentation}. This model first reduces the size of the image and then generates a mask by segmenting the surface water body and assigning an RGB value of 0 to non-water regions. This pre-processing step ensures that subsequent CNN models focus exclusively on water areas, improving the accuracy of optically active parameter predictions. However, the pre-trained model has certain limitations, including its tendency to misclassify sky regions as surface water.

\subsection{Filtering}
Following the segmentation step (Filter \#1) applied to the segmented surface water bodies, an image metadata file is generated containing the columns \texttt{Image\_Path}, \texttt{Site Code}, \texttt{Site Name}, and \texttt{Timestamp}. Using Python's \texttt{astral} library in conjunction with the \texttt{Timestamp} and \texttt{Site Name} columns, the \texttt{State} and \texttt{Day/Night} information for each image is derived. Based on the given \texttt{Timestamp} and \texttt{State}, the \texttt{astral} library retrieved the local timezone for each image.

The \texttt{Timestamp} is converted from UTC to local time with the local timezone information. The \texttt{astral} library is then used to determine each day's sunrise and sunset times at the specified site location in local time. The image capture time is compared against these sunrise and sunset times, all standardized to the local timezone, to classify images as either \texttt{Day} (captured between sunrise and sunset) or \texttt{Night} (captured outside this range). Polar Day and Polar Night conditions are also accounted for in certain locations within the U.S., based on their time zones.

\begin{figure}[htbp]  
    \centering  
    \includegraphics[width=\columnwidth]{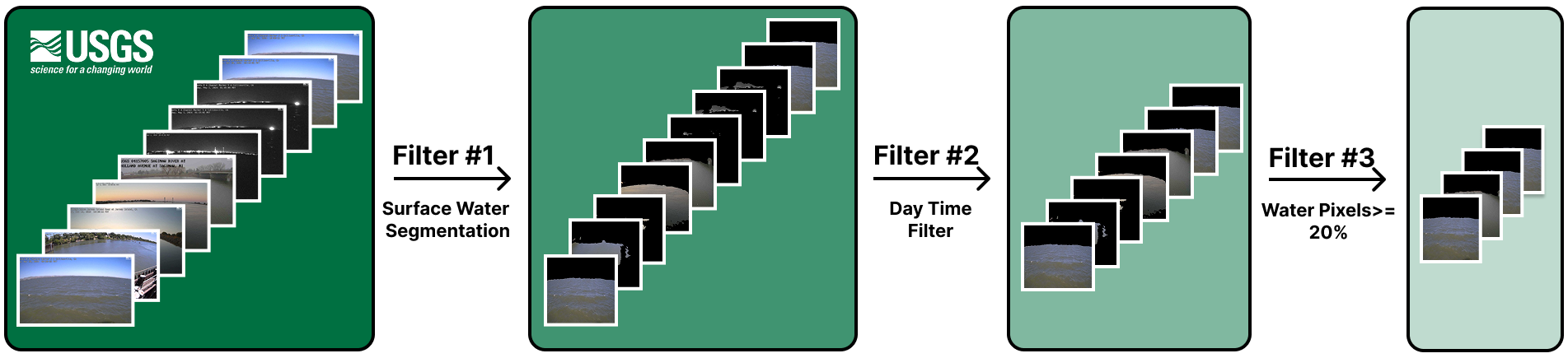}  
    \caption{Reduction in the number of images during the filtering process. The figure illustrates the total number of images before and after applying the segmentation and filtering criteria.}  
    \label{fig:image_filtering}  
\end{figure}

Nighttime images are excluded from the dataset as part of Filter \#2, as low visibility during nighttime hinders the extraction of meaningful information from water body images, rendering them unsuitable for CNN model predictions. This filtering process significantly reduced the number of images retained for the final analysis; the exact reduction in image count is illustrated in Fig.~\ref{fig:image_filtering}.  

The same approach is applied to the variable dataset file containing optically active parameters. This file initially included the columns \texttt{Site Code}, \texttt{Site Name}, \texttt{Timestamp}, and 16 feature columns corresponding to the optically active parameters described in Section~\ref{subsec:DataCollect}. Using Python's \texttt{astral} library along with the \texttt{Timestamp} and \texttt{Site Name} columns, the \texttt{State} and \texttt{Day/Night} information is added to the image's metadata file. Daytime to Nighttime data from various site locations are added to ensure that only reliable daytime data are retained for analysis.

After applying Filter \#2 (daytime filter), an additional filtering step (Filter \#3) is performed on all daytime-classified images. The pre-trained U-Net model had limitations in segmenting surface water under certain conditions, such as varying lighting, challenging site environments, and the absence of ground truth masks. To address this, the segmented surface water area (i.e., the colored pixel region) is computed as a percentage of the total image pixels. To ensure that only relevant data for accurate predictions are retained, images with a segmented surface water area of at least 20\% are kept. These selected images are stored in a dedicated folder, with their metadata recorded in a file. Images failing to meet this criterion are excluded from further analysis.

\subsection{Data Dictionary Design}
After filtering both the image dataset and the variable dataset, the two are merged using the \texttt{Site Code}, \texttt{State}, and \texttt{Day/Night} columns, ensuring that the values in these fields matched exactly, particularly for rows where \texttt{Day/Night = Day}. Additionally, the \texttt{Timestamp} column is used to match each image with the closest recorded parameter measurement. This is achieved using the \texttt{pd.merge\_asof} function from Python's \texttt{pandas} library. Before merging, the \texttt{Timestamp} values in the image dataset are converted to the same format as those in the parameter dataset, aligning from UTC to UTC with the appropriate offset.

After generating the merged dataset, further analysis is conducted to optimize it for CNN models. Among the 16 optically active parameters, each with varying measurement units, only one measurement with the maximum number of non-null values is selected for each parameter. All unnecessary fields are removed, leaving only the relevant attributes: \texttt{Image\_Path}, \texttt{Chlorophyll-$\alpha$}, \texttt{Chlorophylls}, \texttt{CDOM}, \texttt{Phycocyanins}, \texttt{Suspended Sediments}, and \texttt{Turbidity}. Once the relevant subset is finalized, the count of non-null and non-negative values is calculated for each optically active parameter. For each parameter, the valid records are extracted and stored separately, preserving the corresponding \texttt{Image\_Path} and associated values.

This step is carried out to train six individual CNN models, each tailored to predict a specific optically active parameter. By training separate models, computational efficiency is significantly improved, and prediction accuracy is enhanced by focusing on a single parameter at a time.

\begin{figure*}[htbp]
    \centering
    \includegraphics[width=\textwidth]{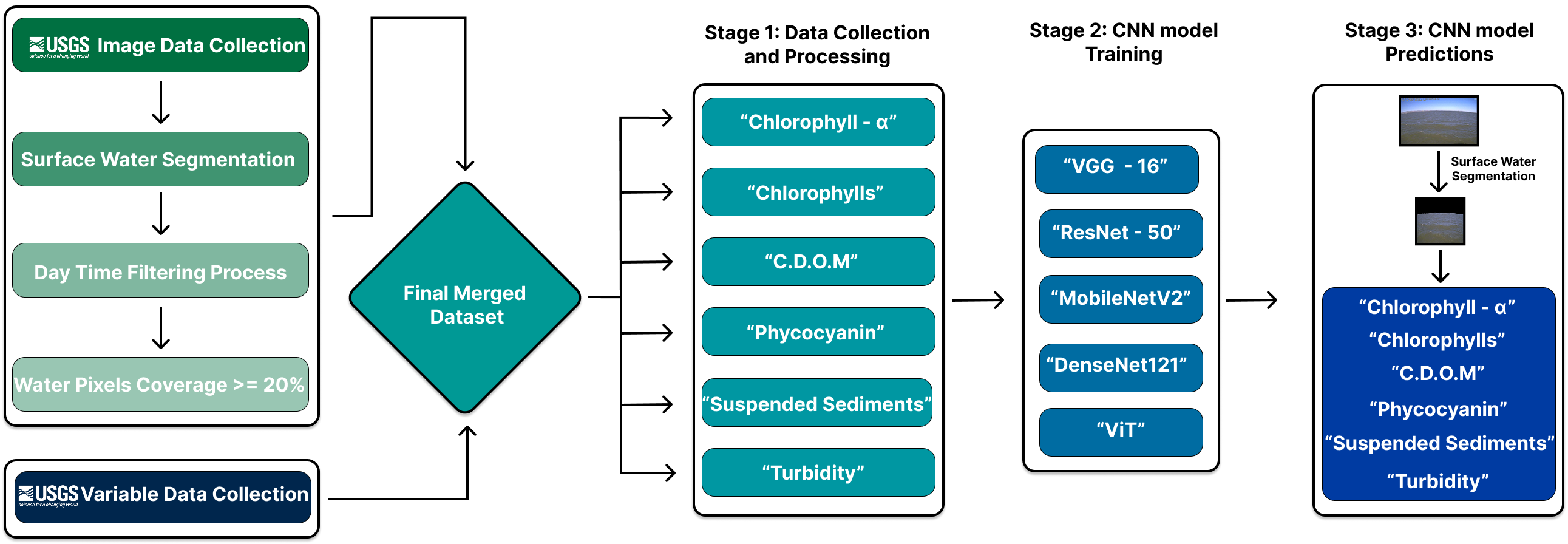}
    \caption{Overview of the data processing and CNN pipeline, including data collection, U-Net segmentation, filtering by pixel thresholds and daytime classification, dataset merging, and model training and prediction.}
    \label{fig:data_pipeline}
\end{figure*}

Table~\ref{table:final_image_count} summarizes the final number of images (or records) retained for each optically active parameter. The resulting labeled datasets served as the structured input, or data dictionary, for HydroVision, enabling the prediction of six optically active parameters from a given image. The complete pipeline, encompassing data collection, segmentation, filtering, merging, model training, and prediction, is illustrated in Fig.~\ref{fig:data_pipeline}.

\begin{table}[ht]
\centering
\caption{Final number of images retained for each optically active parameter with its units.}
\label{table:final_image_count}
\begin{tabular}{|l|r|}
\hline
\textbf{Optically Active Parameter} & \textbf{Number of Images} \\ \hline
Chlorophyll-$\alpha$ (RFU)           & 28,061                            \\ \hline
Chlorophylls (mg/L)                 & 12,681                            \\ \hline
CDOM (ppb QSE)                      & 17,328                            \\ \hline
Phycocyanins (RFU)                  & 18,785                            \\ \hline
Suspended Sediments (mg/L)          & 21,535                            \\ \hline
Turbidity (FNU)                     & 480,125                           \\ \hline
\end{tabular}
\end{table}

\section{Methods}
\label{sec: Methods}
This section details the architecture and design principles of the scene classification models developed for this study. We also describe the architectural choices, including the types of Convolutional Neural Networks employed and the rationale behind their selection, which are important to optimize performance under varied environmental conditions and data constraints.

This decision is driven by a strategic need to address distinct aspects of water quality as discussed in Section~\ref{sec:RelatedWork}, which highlighted the importance of separately analyzing various optically active parameters. By segmenting the classification task into six focused models, we effectively utilized the sparse labeled data available. Each model is tailored and optimized for its specific target variable, ensuring more precise and reliable predictions despite the overarching challenges of data scarcity. This approach also allowed for a more detailed exploration and understanding of individual water quality parameters under diverse environmental conditions.

\subsection{Segmented Images Validation}

To validate the accuracy of the pre-trained U-Net model used for surface water segmentation on the collected dataset, we employed computer vision techniques to create masks distinguishing water and non-water regions in the original images. These masks are compared against the segmented water regions generated by the U-Net model. Since ground truth masks for surface water segmentation are unavailable in our dataset, we utilized computer vision tools to generate surrogate masks for validation purposes.

Various Computer Vision and Machine Learning techniques are explored to generate masks for the original images, including Normalized Difference Water Index (NDWI), histogram thresholding, K-means clustering, Gaussian Mixture Model (GMM) clustering, Gray Level Co-occurrence Matrix (GLCM) texture analysis, and Canny edge detection with largest contour selection. After evaluating these techniques on a subset of images from diverse locations, GMM thresholding is identified as the most effective method for accurately masking water regions in the original images.

GMM thresholding is a probabilistic clustering technique based on Gaussian Mixture Models, which assumes that the data can be represented as a mixture of multiple Gaussian distributions. Each pixel intensity is evaluated for its likelihood of belonging to one of the Gaussian components, enabling the method to adaptively segment water and non-water regions. This adaptability makes GMM thresholding particularly effective in handling variations in lighting, water color, and image quality across the dataset \cite{reynolds2009gaussian}. 

GMM thresholding is subsequently applied to all images that passed the following filtering criteria:
\begin{itemize}
    \item Successfully segmented using the U-Net model,
    \item Captured during daytime (after day/night filtering),
    \item Contained at least 20\% segmented surface water area relative to the total image area (in terms of pixels).
\end{itemize}

The validated segmented images serve as input for CNN-based regression models, enabling prediction of optically active water quality parameters. The next section details the model architectures and training process.

\subsection{HydroVision Architecture}
The final six refined files, combined with image metadata info and the variable data, each corresponding to an optically active parameter, are used to train independent convolutional neural network models. Four pre-trained convolutional neural networks are evaluated: VGG16\cite{simonyan2014very}, ResNet50 \cite{he2016deep}, MobileNetV2\cite{sandler2018mobilenetv2}, and DenseNet121\cite{huang2017densely}, selecting the best-performing architecture for deployment.

VGG16 is a widely used CNN architecture model known for its simplicity and effectiveness in feature extraction, making it well-suited for image-based regression tasks. ResNet50 introduces residual connections, which help mitigate vanishing gradient issues and improve model depth, enabling better generalization. MobileNetV2 is a lightweight architecture optimized for efficiency, making it ideal for deployment in resource-constrained environments while maintaining competitive accuracy. DenseNet121 enhances feature reuse by establishing dense connections between layers, leading to improved information flow and reduced parameter redundancy. These architectures are chosen based on their established performance in computer vision tasks and their ability to extract meaningful spatial patterns from RGB water images.

Each model is initialized with ImageNet weights \cite{imagenet2021update} and modified with a custom regression head, consisting of a Global Average Pooling layer, a dense layer with 512 or 1024 neurons, ReLU activation, and L2 regularization to mitigate overfitting. The final output layer is a single neuron with a linear activation function, predicting a continuous-valued optically active parameter. To improve feature adaptation, the last 10 layers of each CNN are unfrozen for fine-tuning, enabling the models to refine high-level representations while retaining lower-level features. The dataset is split into 80\% training and 20\% validation, with batch sizes optimized using Bayesian hyperparameter optimization. Images are resized to 224$\times$224 pixels, normalized using model-specific pre-processing functions, and augmented with random horizontal flipping. The training pipeline used the TensorFlow Dataset API for efficient image loading, caching, shuffling, and prefetching.

For the five optically active parameters (Chlorophyll-$\alpha$, Chlorophylls, CDOM, Phycocyanins, and Suspended Sediments), training is conducted using Bayesian optimization to fine-tune hyperparameters. Bayesian optimization follows a probabilistic approach, leveraging past evaluations to focus on the most promising regions of the hyperparameter space. The optimizer explored the following hyperparameter ranges:

\begin{itemize}
    \item Dropout Rate: 0.3--0.5
    \item L2 Regularization Strength: $1\mathrm{e}{-4}$ to $1\mathrm{e}{-2}$
    \item Learning Rate: $1e^{-5}$ to $1e^{-3}$
    \item Number of Neurons in Dense Layer: 512 or 1024
    \item Optimizer Selection: Adam or Stochastic Gradient Descent (SGD)	
\end{itemize}

The training is carried out for a maximum of 50 epochs, with early stopping applied if the validation loss plateaus for five consecutive epochs. Additional training strategies included:

\begin{itemize}
    \item Model check-pointing to save the best-performing model based on validation loss.
    \item Dynamic learning rate adjustment using \texttt{ReduceLROnPlateau} with a patience of two epochs.
\end{itemize}

To enhance computational efficiency, training is performed on a GPU-accelerated environment with memory growth enabled and mixed precision training where supported. Model performance is evaluated using multiple metrics, including Mean Squared Error (MSE), Mean Absolute Error (MAE), Root Mean Squared Error (RMSE), R-squared (R²), and Symmetric Mean Absolute Percentage Error (sMAPE), guiding the selection of the most effective architecture.

The final optically active parameter, Turbidity, is trained separately due to its significantly larger dataset size (480,125 images) compared to the other parameters. To optimize GPU memory usage and reduce training overhead, the following modifications are applied:

\begin{itemize}
    \item Enabled TensorFlow XLA Compilation to improve computational efficiency.
    \item Used a batch size of 8, balancing memory constraints and training stability.
    \item Applied a Cosine Decay Learning Rate Scheduler to dynamically adjust the learning rate during training.
    \item Unfroze the last 10 layers of the CNN models for fine-tuning, allowing the model to adapt high-level features to the task.
    \item Applied a stricter \texttt{ReduceLROnPlateau} schedule, reducing the learning rate after just one epoch of stagnation instead of two.
\end{itemize}

Unlike the other five parameters, Bayesian optimization is not used for Turbidity. Instead, fixed hyperparameters are selected based on prior experiments. Additionally, random horizontal flipping is not applied to Turbidity due to its dataset characteristics.

The Turbidity dataset followed the same pre-processing and augmentation pipeline as the other five parameters, with images resized to 224$\times$224 pixels and normalized using the selected CNN model’s pre-processing function.

In addition to CNNs, we evaluated a ViT model~\cite {dosovitskiy2020image} for Suspended Sediments and Turbidity, using the \texttt{vit-base-patch16-224}~\cite{wu2020visual} architecture pre-trained on ImageNet\cite{imagenet2021update}. The model is adapted for regression by extracting the [CLS] token, followed by a dropout layer, a dense ReLU layer, and a linear output. Inputs are resized to 224$\times$224, pre-processed using Hugging Face’s \texttt{ViTImageProcessor}, and passed in channel-first format. The last three transformer blocks are unfrozen for fine-tuning. Suspended Sediments used the same Bayesian optimization strategy as CNNs, while Turbidity is trained with fixed hyperparameters for 300 epochs using cosine learning rate decay and early stopping.

\section{Results}
\label{sec:Results}
This section presents the results of surface water segmentation accuracy used during data collection, as well as the performance of the CNN models trained for optically active parameter estimation. The results help justify methodological choices made throughout the project and demonstrate the effectiveness of our proposed approach.

\subsection{Segmentation Results}
\begin{figure*}[h]
    \centering
    \includegraphics[width=\linewidth]{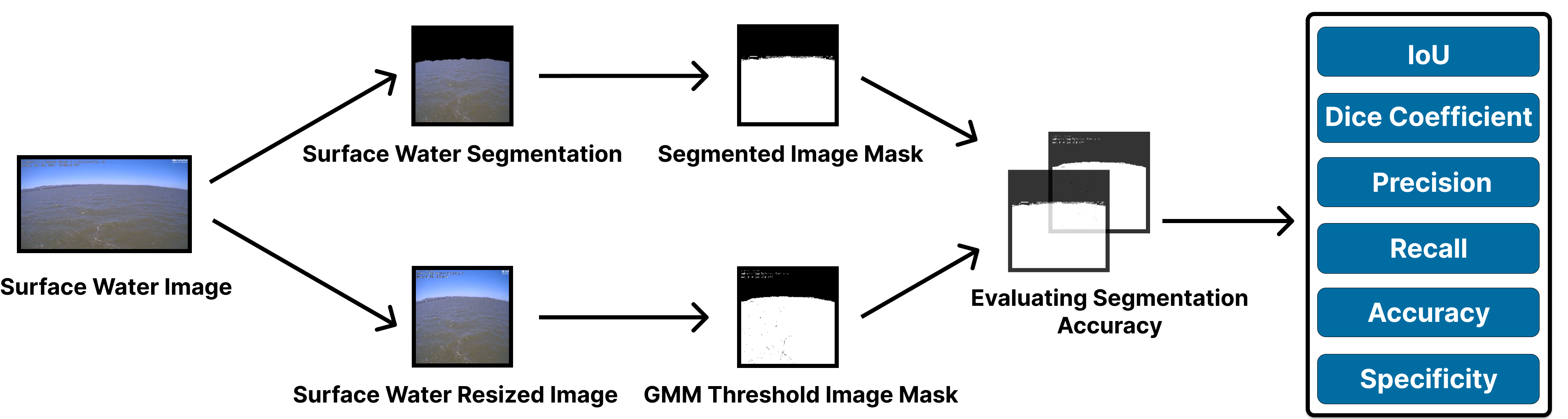}
    \caption{Segmentation accuracy evaluation pipeline using U-Net and GMM masks, with metrics including IoU, Dice, Precision, Recall, Accuracy, and Specificity.}
    \label{fig:segmentation_evaluation_workflow}
\end{figure*}

To evaluate the performance of the U-Net segmentation model, we considered key segmentation metrics, including Intersection over Union (IoU), Dice coefficient, Precision, Recall, Accuracy, and Specificity. To ensure that only high-quality segmentations are analyzed, we applied filtering thresholds of IoU \textgreater 0.5, Dice \textgreater 0.2, Precision \textgreater 0.1, and Recall \textgreater 0.2, removing low-confidence segmentations from the dataset. Fig.~\ref{fig:segmentation_evaluation_workflow} illustrates the pipeline used to evaluate segmentation accuracy by comparing U-Net-generated masks with GMM-based reference masks.

Table~\ref{tab:filtered_segmentation_metrics} summarizes the performance of the segmentation model after filtering. The mean IoU is 0.524, indicating moderate alignment between the segmented and reference water regions. The mean Dice coefficient of 0.687 suggests that the segmented regions exhibit strong overlap with the ground truth. Notably, the recall value of 0.949 indicates that most water pixels are correctly identified, though the specificity (0.472) suggests occasional misclassification of non-water regions as water.

\begin{table}[h]
    \centering
    \caption{Summary of Segmentation Metrics for Filtered Data (IoU \textgreater 0.5, Dice \textgreater 0.2, Precision \textgreater 0.1, Recall \textgreater 0.2)}
    \label{tab:filtered_segmentation_metrics}
    \begin{tabular}{|l|c|c|}
        \hline
        \textbf{Metric} & \textbf{Mean} & \textbf{Standard Deviation} \\
        \hline
        IoU & 0.524 & 0.017 \\
        \hline
        Dice Coefficient & 0.687 & 0.014 \\
        \hline
        Precision & 0.541 & 0.033 \\
        \hline
        Recall & 0.949 & 0.050 \\
        \hline
        Accuracy & 0.670 & 0.071 \\
        \hline
        Specificity & 0.472 & 0.178 \\
        \hline
    \end{tabular}
\end{table}

The segmentation model effectively identifies water regions, with high recall ensuring most water pixels are detected. However, moderate IoU and low specificity indicate occasional misclassification of non-water regions due to reflections or sky-like colors. Refining the training dataset and applying post-processing techniques, such as manual masking, can help improve accuracy. Given U-Net’s limitations, an additional filter is applied: only segmented images with surface water pixels covering at least 20\% of the image are retained. This threshold is empirically chosen after evaluating multiple coverage levels (e.g., 25\%, 30\%, 35\%, 40\% and 50\%). The 20\% cutoff is found to strike the best balance between dataset size and segmentation reliability—preserving enough diverse training samples while ensuring spatially meaningful water coverage. This threshold offered an optimal trade-off between training volume and quality for robust model performance.

While the filtered mean IoU (0.524) is moderate, it is relatively low compared to typical binary segmentation tasks. This is largely attributed to the challenge of distinguishing water from sky and reflective surfaces in RGB imagery, where color and texture can be visually ambiguous. These issues are especially pronounced under overcast lighting, low solar angles, or high glare conditions—leading to occasional misclassification of sky regions as water. Future work should incorporate class-weighted segmentation metrics, including per-class IoU and Dice scores, to more explicitly quantify errors for both water and non-water regions. Additionally, methods such as horizon line detection, sun-glint correction, or attention-based segmentation could improve performance by reducing false positives caused by sky-colored regions.

\subsection{HydroVision Results}

This section evaluates the performance of four CNN architectures—VGG16, ResNet50, MobileNetV2, DenseNet121, and ViT trained for predicting optically active water quality parameters. The models are assessed using multiple regression-based evaluation metrics, including mean squared error (MSE), mean absolute error (MAE), root mean squared error (RMSE), validation R-squared (Val $R^2$), and symmetric mean absolute percentage error (sMAPE). These metrics provide a comprehensive assessment of model accuracy, robustness, and generalization.

Table~\ref{tab:model_performance} summarizes the final training and validation results for each model. Performance is evaluated on the test dataset described in Section~\ref{sec: Methods}, with a focus on generalizability across different water quality parameters.

\begin{table}[h]
    \centering
    \caption{Performance comparison of different CNN models for optically active water quality parameter estimation.}
    \label{tab:model_performance}
    \resizebox{\columnwidth}{!}{%
    \begin{tabular}{|l|c|c|c|c|c|c|}
        \hline
        \textbf{Parameter} & \textbf{Model} & \textbf{MSE} & \textbf{MAE} & \textbf{RMSE} & \textbf{Val R²} & \textbf{sMAPE} \\
        \hline
        CDOM                 & VGG16       & 67.66  & 5.21   & 8.23   & 0.737   & 28.70   \\
                             & ResNet50    & 31.59  & 3.39   & 5.62   & 0.874   & 19.52   \\
                             & MobileNetV2 & 38.42  & 3.78   & 6.20   & 0.847   & 21.43   \\
                             & \textbf{DenseNet121} & 25.04  & 2.99   & 5.00   & \textbf{0.898}   & 18.43   \\
        \hline
        Chlorophyll-$\alpha$ & VGG16       & 2.38   & 0.64   & 1.54   & 0.424   & 52.69   \\
                             & ResNet50    & 1.45   & 0.44   & 1.21   & 0.631   & 45.76   \\
                             & MobileNetV2 & 1.46   & 0.45   & 1.21   & 0.606   & 46.08   \\
                             & \textbf{DenseNet121} & 1.32   & 0.40   & 1.14   & \textbf{0.678}   & 39.24   \\
        \hline
        Chlorophylls         & VGG16       & 52.50  & 4.79   & 7.25   & 0.645   & 74.73   \\
                             & ResNet50    & 42.08  & 3.66   & 6.49   & 0.712   & 52.63   \\
                             & MobileNetV2 & 38.71  & 3.55   & 6.22   & 0.732   & 51.08   \\
                             & \textbf{DenseNet121} & 30.06  & 3.17   & 5.48   & \textbf{0.788}   & 49.53   \\
        \hline
        Phycocyanin          & VGG16       & 4.81   & 0.96   & 2.19   & 0.567   & 72.36   \\
                             & ResNet50    & 4.88   & 1.06   & 2.21   & 0.583   & 95.93   \\
                             & MobileNetV2 & 3.61   & 0.75   & 1.90   & 0.693   & 61.02   \\
                             & \textbf{DenseNet121} & 2.67   & 0.64   & 1.63   & \textbf{0.779}   & 50.77   \\
        \hline
        Suspended Sediments  & VGG16       & 16682.24 & 19.18 & 129.16 & NaN  & 52.45   \\
                             & ResNet50    & 13523.04 & 18.54 & 116.29 & NaN  & 47.16   \\
                             & \textbf{MobileNetV2} & 17423.69 & 19.99 & 131.99 & NaN  & 49.79   \\
                             & DenseNet121 & 17356.99 & 18.83 & 131.75 & NaN  & 41.75   \\
                             & ViT         & 0.322   & 0.1201 & 0.5672 & NaN  & 40.28   \\
        \hline
        Turbidity            & VGG16       & 22099.57 & 37.94 & 148.36 & NaN  & 87.46   \\
                             & ResNet50    & 13923.09 & 26.50 & 117.77 & 0.031   & 70.46   \\
                             & MobileNetV2 & 16233.90 & 27.81 & 127.28 & 0.185   & 71.05   \\
                             & \textbf{DenseNet121} & 16746.22 & 22.66 & 129.35 & \textbf{0.498}   & 58.02   \\
                             & ViT         & 0.5653 & 0.225 & 0.752 & NaN & 107.77 \\
        \hline
    \end{tabular}
    }
\end{table}

DenseNet121 achieved the best overall performance, particularly for CDOM (Val R² = 0.898) and Chlorophylls (Val R² = 0.788), likely due to its dense connectivity, which promotes feature reuse and enhances gradient flow, making it more effective at capturing subtle spatial patterns in water imagery. ResNet50 also performed well for CDOM (Val R² = 0.874) but showed inconsistent results across other parameters, possibly due to its deeper architecture being more sensitive to noise and overfitting on limited samples. MobileNetV2 balanced performance and efficiency, benefiting from depthwise separable convolutions, but lacked the representational power needed for more complex parameters. VGG16 consistently underperformed—particularly for Chlorophyll-$\alpha$ (Val R² = 0.424) and Phycocyanin (Val R² = 0.567), highlighting its limited capacity for hierarchical feature extraction due to its shallow depth and absence of skip connections. For Suspended Sediments, all CNN models yielded negative validation R² scores, reflecting the challenge of modeling dynamic sediment patterns from RGB imagery alone. Nonetheless, DenseNet121 showed relative robustness with the lowest MSE (17,356.99) and best sMAPE (41.75). The high RMSE values across all models suggest that multispectral or temporal information may be necessary to improve prediction accuracy. For Turbidity, DenseNet121 again performed best (Val R² = 0.498), benefiting from its ability to capture fine-grained patterns and high-frequency variations, while ResNet50 (Val R² = 0.030) and MobileNetV2 (Val R² = 0.185) showed only modest generalization. VGG16 performed the worst (Val R² = -0.162), likely due to its inability to distinguish turbidity-related features from environmental artifacts such as glare, reflections, or cloud cover.

We additionally evaluate the ViT architecture for Suspended Sediments using the \texttt{vit-base-patch16-224} architecture. While ViT achieved the lowest RMSE (0.5672) and sMAPE (40.28), its negative validation R² (-4.91) suggests limited generalization, likely due to its data-hungry nature and sensitivity to noise. For Turbidity, ViT is again evaluated using the \texttt{vit-base-patch16-224} architecture. Despite achieving the lowest RMSE (0.752) and MAE (0.225) among all models, it resulted in a significantly negative validation $R^2$ score of -83.62, indicating poor generalization due to a mismatch with the validation data distribution. The high sMAPE (107.77) reinforces this, indicating large relative errors in prediction—possibly due to the model's sensitivity to noise and high data variance in Turbidity samples.

Negative validation $R^2$ values observed in Table~\ref{tab:model_performance}, particularly for Suspended Sediments and Turbidity, signify that the models performed worse than a naive baseline regressor that simply predicts the mean of the target variable. An $R^2$ value less than zero implies that the residual sum of squares from the model predictions exceeds the total variance in the target variable. This under performance was especially evident for VGG16 (Val $R^2 = -0.162$) and ViT (Val $R^2 = -83.62$), which failed to generalize to the validation data despite achieving low RMSE or MAE on the training set. ViT's sensitivity to noise and data imbalance, in particular, led to extreme overfitting—reinforcing its unsuitability without larger and more diverse training samples. This is a strong indicator of model failure in capturing the underlying data distribution. In the context of our task, this underperformance is likely attributed to several factors. Suspended Sediments and Turbidity are known to exhibit high variability in natural water bodies due to the influence of dynamic environmental conditions such as wave patterns, surface glare, floating debris, seasonal runoff, and lighting variations \cite{w16020306}. RGB imagery is particularly limited in capturing these complex, transient phenomena, as it lacks the spectral sensitivity needed to isolate key sediment-related features \cite{app13127219}. Moreover, water surface reflections and atmospheric distortions (e.g., clouds, shadows) introduce additional noise that can mislead CNNs during feature learning \cite{9711422}. In the context of our task, this under performance is likely attributed to several factors. Suspended Sediments and Turbidity are known to exhibit high variability in natural water bodies due to the influence of dynamic environmental conditions such as wave patterns, surface glare, floating debris, seasonal runoff, and lighting variations \cite{w16020306}. RGB imagery is particularly limited in capturing these complex, transient phenomena, as it lacks the spectral sensitivity needed to isolate key sediment-related features \cite{app13127219}. Moreover, water surface reflections and atmospheric distortions (e.g., clouds, shadows) introduce additional noise that can mislead CNNs during feature learning \cite{9711422}.

Another contributing factor could be the imbalanced distribution of values in the training set, where certain ranges (e.g., low turbidity or sediment levels) are overrepresented. This may bias the models toward regressing to a narrow central tendency and fail to generalize across the full range of target values \cite{bdcc7010015}. While DenseNet121 and MobileNetV2 still exhibited negative $R^2$ values, they demonstrated relatively better robustness than other models by achieving lower RMSE and sMAPE scores—suggesting closer average predictions despite poor overall generalization. This underscores the difficulty of learning from RGB data alone without additional spectral (e.g., NIR) or temporal information. These findings emphasize the importance of dataset diversity and advanced feature extraction techniques for modeling optically active but visually ambiguous parameters.

To evaluate HydroVision in real-world environments, we collect a validation dataset from November 5, 2024, to February 17, 2025, using the same data collection, filtering, and pre-processing steps. The dataset is used to assess the best-trained models for each CNN architecture and optically active parameter. Table~\ref{tab:validation_results} presents the results. Rather than relying solely on traditional metrics like $R^2$ or sMAPE—which can be sensitive to outliers, extreme values, or non-normal distributions common in environmental data—we incorporate a set of metrics better aligned with real-world applicability and interpretability \cite{10.2166/nh.2021.071, MCKENZIE2011259}. These include: Pearson's Correlation Coefficient (r), which measures the strength and direction of the linear relationship between predictions and actual values, where a strong positive correlation (close to 1) indicates the model captures meaningful trends; and Concordance Correlation Coefficient (CCC), which accounts for both precision and accuracy in regression tasks involving continuous variables.

%%%%%%%%%%%%%%%%%%%%%%%%%%%%%%%%
\begin{table}[h]
    \centering
    \caption{Validation dataset performance of best models for each optically active parameter.}
    \label{tab:validation_results}
    \resizebox{\columnwidth}{!}{%
    \begin{tabular}{|l|l|c|c|}
        \hline
        \textbf{Parameter} & \textbf{Model} & \textbf{Pearson r} & \textbf{CCC} \\
        \hline
        CDOM                 & VGG16       & 0.589  & 0.304  \\
                             & \textbf{ResNet50}    & \textbf{0.792}  & 0.707  \\
                             & MobileNetV2 & 0.753  & 0.632  \\
                             & DenseNet121 & 0.736  & 0.626  \\
        \hline
        Chlorophyll-$\alpha$ & VGG16       & 0.350  & 0.340  \\
                             & ResNet50    & 0.370  & 0.353  \\
                             & \textbf{MobileNetV2} & \textbf{0.404}  & 0.347  \\
                             & DenseNet121 & 0.395  & 0.364  \\
        \hline
        Chlorophylls         & VGG16       & 0.129  & 0.091  \\
                             & ResNet50    & NaN & NaN \\
                             & MobileNetV2 & NaN & NaN \\
                             & \textbf{DenseNet121} & \textbf{0.006}  & 0.001  \\
        \hline
        Phycocyanin          & \textbf{VGG16}       & \textbf{0.150}  & 0.095  \\
                             & ResNet50    & 0.105  & 0.075  \\
                             & MobileNetV2 & 0.073  & 0.045  \\
                             & DenseNet121 & 0.056  & 0.034  \\
        \hline
        Suspended Sediments  & VGG16       & 0.113  & 0.018  \\
                             & ResNet50    & 0.056  & 0.036  \\
                             & \textbf{MobileNetV2} & \textbf{0.187}  & 0.058  \\
                             & DenseNet121 & 0.073  & 0.045  \\
                             & ViT         & 0.042  & 0.000  \\
        \hline
        Turbidity            & VGG16       & NaN     & 0.000  \\
                             & ResNet50    & 0.363  & 0.000  \\
                             & \textbf{MobileNetV2} & \textbf{0.418}  & 0.000  \\
                             & DenseNet121 & 0.410  & 0.000  \\
                             & ViT         & NaN & NaN \\
        \hline
    \end{tabular}%
    }
\end{table}

The validation results show that ResNet50 and MobileNetV2 performed best for CDOM and Chlorophyll-$\alpha$, respectively, with Pearson correlation coefficient $r$ values of 0.792 and 0.404. ResNet50 also achieved the highest CCC (0.707) for CDOM, reflecting strong predictive alignment. Their superior performance is likely due to residual connections in ResNet50 and the efficient depthwise separable convolutions in MobileNetV2, both enabling better feature extraction. VGG16 and DenseNet121 underperformed slightly for CDOM, likely due to limited depth or reduced generalization. For Chlorophylls, all models showed weak correlation (\( r \approx 0 \)), with DenseNet121 performing marginally better ($r = 0.006$), indicating difficulty in distinguishing overlapping spectral patterns. VGG16 yielded the highest correlation for Phycocyanin ($r = 0.150$), though all models struggled to learn meaningful representations (\( r < 0.15 \)). Similarly, MobileNetV2  gave the highest correlation for Suspended Sediments ($r = 0.187$), although all models showed poor agreement, suggesting systematic underestimation. For Turbidity, MobileNetV2 also achieved the highest Pearson correlation coefficient ($r = 0.418$), followed closely by DenseNet121 ($r = 0.410$), though CCC values remained low across all models. VGG16 yielded constant predictions across all samples, producing the same value regardless of input—which led to a standard deviation near zero and an undefined Pearson correlation coefficient $r$ value (NaN). This suggests that the model failed to learn meaningful turbidity-specific features, possibly due to over-regularization or poor convergence.

In Table~\ref{tab:validation_results}, we observe several cases of low or even negative Pearson's $r$ and Concordance Correlation Coefficient (CCC), particularly for Chlorophylls, Phycocyanin, Suspended Sediments, and Turbidity. These metrics quantify the linear correlation and agreement between predicted and actual values, and their degradation suggests that model outputs not only deviate significantly from ground truth but also fail to follow the same directional trends. Such weak correlations imply that models struggled to capture meaningful patterns during training and were unable to transfer learned features to unseen validation samples.

The issue is most prominent for the VGG16 model on Turbidity, where Pearson’s $r$ is marked as NaN (Not a Number). This occurred because the model predicted a constant value across all validation samples, resulting in zero variance in its predictions. Pearson’s correlation is undefined when either variable has zero standard deviation, as the denominator in the correlation formula becomes zero. A closer look at the VGG16 prediction distribution confirms this, with minimum and maximum predicted values being identical and standard deviation virtually zero. Such behavior suggests that the model either converged to a trivial constant output during training or suffered from vanishing gradients or over-regularization, leading it to collapse into a degenerate solution. More generally, negative correlation and agreement values may also be attributed to the spectral ambiguity of the water parameters in RGB images. Phycocyanin and Chlorophylls, for example, have weak signatures in the visible spectrum and are more effectively captured in near-infrared or hyperspectral bands. The lack of distinctive features, coupled with high noise and variation in lighting or turbidity conditions, may cause CNNs to overfit to superficial textures or water color hues that are not generalizable. This misalignment becomes evident during validation, where unseen conditions expose the fragility of the learned representations. These observations underscore the need for additional regularization, spectral data augmentation, or multimodal inputs in future iterations of HydroVision. Leveraging ancillary data such as satellite-derived turbidity indices or incorporating temporal trends may also aid in better modeling parameters that exhibit low visual discriminability in RGB-only datasets.

\begin{figure}[!t]
\centering
% Row 1
\subfloat[CDOM]{\includegraphics[width=0.22\textwidth]{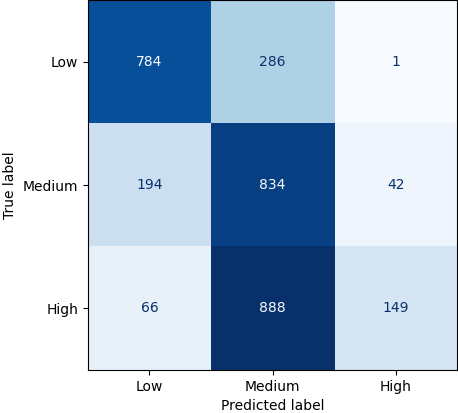}}\hfil
\subfloat[Chlorophyll-$\alpha$]{\includegraphics[width=0.22\textwidth]{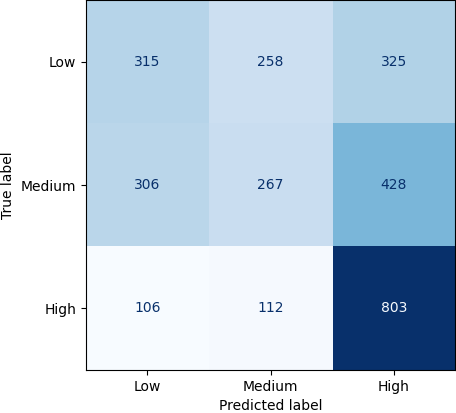}}\\[0.5ex]

% Row 2
\subfloat[Chlorophylls]{\includegraphics[width=0.22\textwidth]{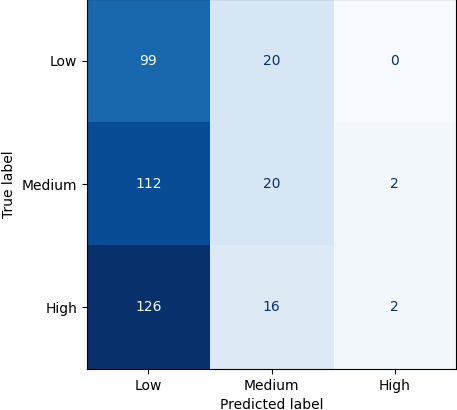}}\hfil
\subfloat[Phycocyanin]{\includegraphics[width=0.22\textwidth]{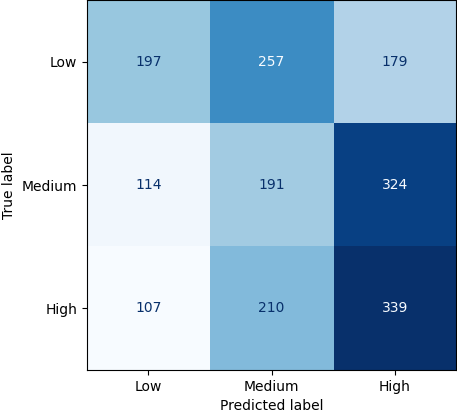}}\\[0.5ex]

% Row 3
\subfloat[Suspended\\Sediments]{\includegraphics[width=0.22\textwidth]{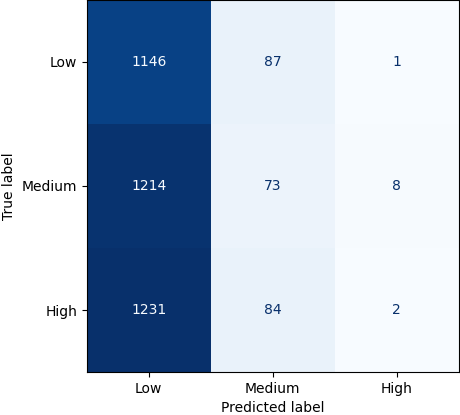}}\hfil
\subfloat[Turbidity]{\includegraphics[width=0.22\textwidth]{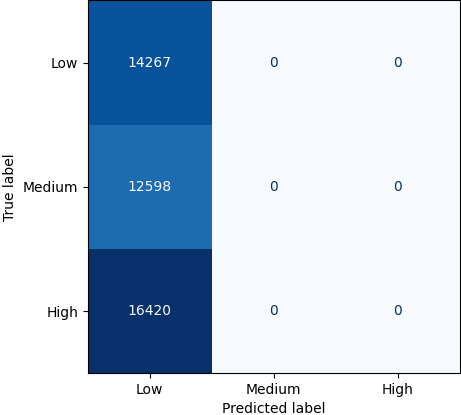}}

\caption{Confusion matrices showing classification-style evaluation of the best model for each optically active parameter. Labels and predictions were binned into Low, Medium, and High categories using quantiles.}
\label{fig:confusion_matrices_all}
\end{figure}

In addition to regression metrics, a classification-style evaluation is performed by discretizing both predicted and actual values of each optically active parameter into three quantile-based bins—\textit{Low}, \textit{Medium}, and \textit{High}. These bins were based on the 33rd and 66th percentiles of ground truth values. Confusion matrices were then generated for the best model of each parameter to visualize how well predictions aligned with the true concentration class. Fig.~\ref{fig:confusion_matrices_all} presents confusion matrices for all six parameters. CNN models used standard Keras pre-processing pipelines, while the ViT model (used for Suspended Sediments and Turbidity) required channels-first input and Hugging Face's \texttt{ViTImageProcessor} for feature extraction. Fig.~\ref{fig:confusion_matrices_all} presents updated confusion matrices based on the best-performing models on the validation dataset.

For CDOM, ResNet50 demonstrated strong classification performance, with clear diagonal dominance and minimal misclassifications, aligning with its high Pearson \( r = 0.7915 \) and CCC. For Chlorophyll-$\alpha$, MobileNetV2 achieved reasonable separation between classes, though some confusion remained between \textit{Low} and \textit{Medium} bins, reflecting the subtlety of spectral differences in this parameter. For Chlorophylls, DenseNet121 showed moderate classification ability, with noticeable misclassifications across all bins, particularly in the \textit{Low} and \textit{Medium} ranges—mirroring the weak correlation (\( r = 0.0055 \)) and low agreement. In Phycocyanin, VGG16 exhibited a strong bias toward the \textit{Low} class with heavy misclassification of \textit{Medium} and \textit{High} values, consistent with limited spectral separability and low performance metrics. For Suspended Sediments, VGG16 struggled to clearly separate classes, showing significant overlap between \textit{Low}, \textit{Medium}, and \textit{High} predictions—highlighting poor agreement despite being the best performer. Finally, for Turbidity, MobileNetV2 provided the most balanced predictions among classes, with better class separation than other models. However, confusion between \textit{Medium} and \textit{High} bins remained, consistent with its modest correlation and low CCC.

Across Chlorophylls, Suspended Sediments, and Turbidity, a broader pattern of model underestimation is evident, with a noticeable bias toward the \textit{Low} prediction class. This likely stems from class imbalance in the training data, where lower concentration levels dominate, and from the limited spectral resolution of RGB imagery, which reduces model sensitivity to high-range values. Such distributional skew can lead models to regress toward the central tendency, under-representing medium and high concentrations in classification outputs.

\begin{figure}[!t]
\centering
% Row 1
\subfloat[CDOM]{\includegraphics[width=0.22\textwidth]{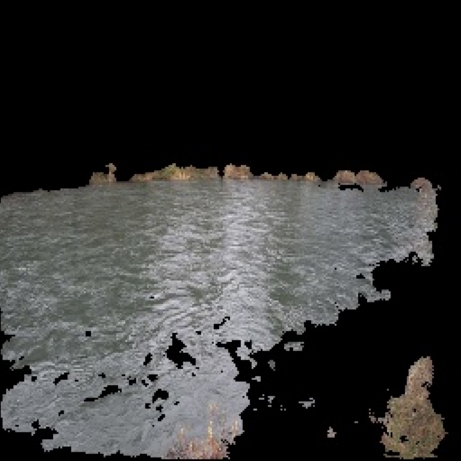}}\hfil
\subfloat[Chlorophyll-$\alpha$]{\includegraphics[width=0.22\textwidth]{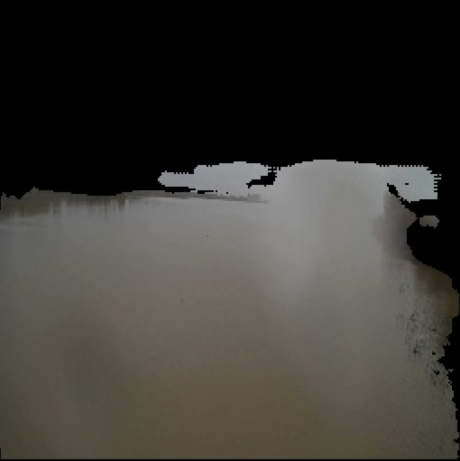}}\\[0.5ex]

% Row 2
\subfloat[Chlorophylls]{\includegraphics[width=0.22\textwidth]{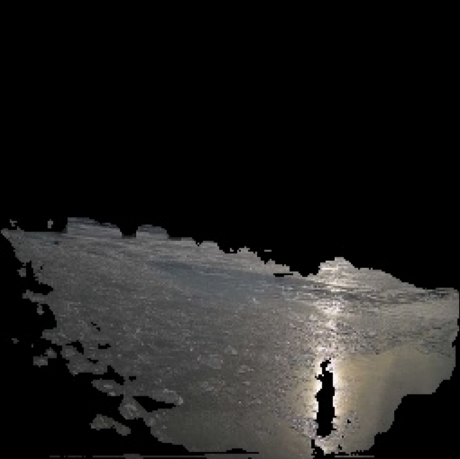}}\hfil
\subfloat[Phycocyanin]{\includegraphics[width=0.22\textwidth]{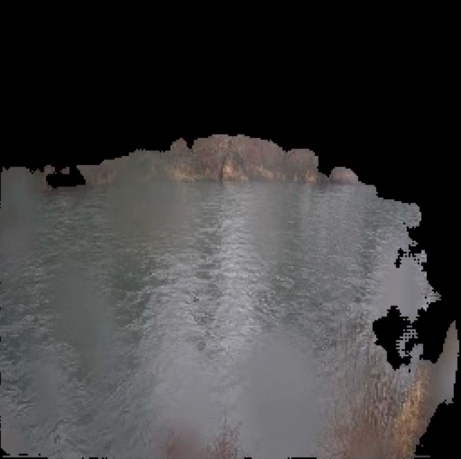}}\\[0.5ex]

% Row 3
\subfloat[Suspended\\Sediments]{\includegraphics[width=0.22\textwidth]{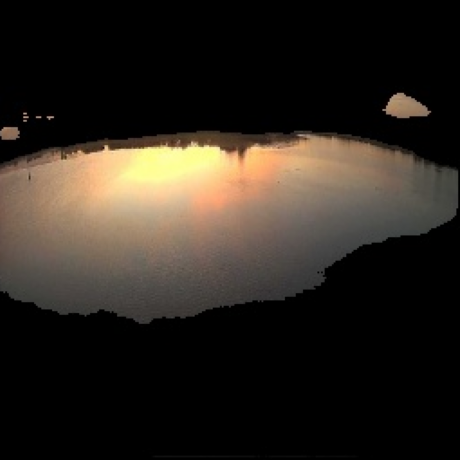}}\hfil
\subfloat[Turbidity]{\includegraphics[width=0.22\textwidth]{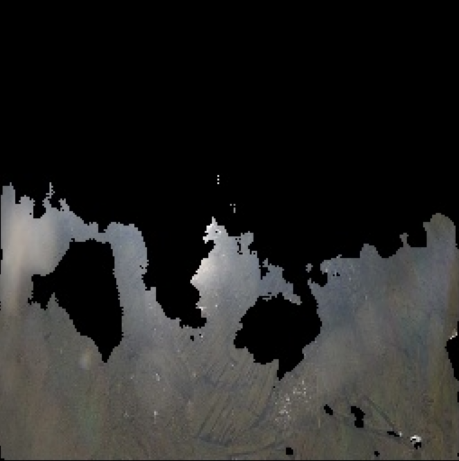}}

\caption{
Representative predictions for each optically active parameter using different CNN models. Each image shows the predicted and actual values, offering qualitative insight into model performance. Subfigures: 
        (a) CDOM, 
        (b) Chlorophyll-$\alpha$, 
        (c) Chlorophylls, 
        (d) Phycocyanin, 
        (e) Suspended Sediments, 
        (f) Turbidity.
}
\label{fig:validation_qualitative_examples}
\end{figure}

To better illustrate these findings, Fig.~\ref{fig:validation_qualitative_examples} presents representative validation examples from different CNN models for each optically active parameter, highlighting how well model predictions align with actual values. Specifically, ResNet50 accurately predicted CDOM (Actual: 14.04, Predicted: 15.22), while MobileNetV2 performed well for Chlorophyll-$\alpha$ (Actual: 0.40, Predicted: 0.45) and Chlorophylls (Actual: 0.20, Predicted: 0.28). VGG16 predicted Phycocyanin with reasonable accuracy (Actual: 0.38, Predicted: 0.58) and showed moderate deviation for Suspended Sediments (Actual: 5.40, Predicted: 8.51). MobileNetV2 underestimated Turbidity (Actual: 1.10, Predicted: 0.01). These examples visually demonstrate the ability of the best-trained models to approximate ground truth values, despite the inherent variability and complexity of natural aquatic scenes.

The evaluation of HydroVision highlights the effectiveness of deep learning models in predicting optically active water quality parameters, with DenseNet121 and ResNet50 consistently delivering strong performance across multiple metrics. ViT showed potential for Suspended Sediments by achieving the lowest RMSE and sMAPE, though its low CCC and high negative bias indicate overfitting and poor generalization. Persistent difficulties have been observed in modeling Phycocyanin, Chlorophylls, and Suspended 
Sediments—likely due to their weak spectral signals in RGB imagery and limited variance in training data. Turbidity has shown mixed outcomes: while CNNs like DenseNet121 and MobileNetV2 achieved modest correlation scores, all models struggled with agreement metrics and exhibited high bias. Notably, VGG16 failed to learn meaningful turbidity-specific features, collapsing to constant predictions that resulted in an undefined Pearson's $r$ (NaN). These results highlight the importance of improved data augmentation, enhanced feature diversity, and architectural tuning—including consideration of spectral or temporal data—to achieve better generalization for visually ambiguous water quality parameters.

\section{Discussions \& Conclusion}
\label{sec:Discussion}
This section discusses aspects of the HydroVision model used in evaluating surface water quality using Computer Vision techniques. We revisit our research questions around (1) identifying labeled datasets for training environmental monitoring scene classification models and (2) evaluating Computer Vision techniques on optically active water quality parameters during seasonal conditions. We also conclude with broader implications and limitations of HydroVision.

\subsection{RQ1: Dataset Sourcing and Preparation for Scene Classification}
As discussed in Section~\ref{sec:RelatedWork}, crucial limitations exist in publicly available, real-world datasets that can influence computer vision model performance. After extensive research and evaluations of accessible datasets, we identified and implemented a carefully designed pre-processing pipeline to enhance the original source, making it ideal for computer vision model development. One of the primary challenges in dataset sourcing is the inconsistency in data availability across monitoring sites, as only a subset of locations collected both imagery and optically active parameters. Additionally, variations in measurement units across different stations introduced complexity in standardizing the data, requiring careful selection and conversion to ensure consistency without compromising scientific integrity.

Beyond sourcing difficulties, the raw image data presented challenges due to environmental variability and scene complexity. Unlike controlled laboratory conditions, real-world images captured by USGS HIVIS cameras included extraneous background elements such as vegetation, infrastructure, and varying weather conditions, which could interfere with model performance. To mitigate this, we employed a pre-trained U-Net segmentation model to isolate water bodies, reducing the influence of non-water features. However, misclassification of sky regions as water remained an issue, requiring additional filtering based on pixel distribution to ensure that only images with a sufficient proportion of visible water are retained. 

Additionally, since night-time images lacked adequate illumination for reliable optical analysis, we incorporated an automated day/night classification method using solar position calculations to exclude low-visibility samples. By integrating these pre-processing steps, we enhanced dataset quality, ensuring that images are relevant to the study and optimized for CNN model training. As such, we are making this data set available \url{https://github.com/AI-VTRC/HydroVision} to the academic community in hopes of further inspiring computer vision research related to surface water monitoring.

\subsection{RQ2: Evaluating AI in Quantifying Water Quality from RGB Images}

The effectiveness of AI-driven computer vision models in estimating optically active water quality parameters from RGB images is the key focus of this study. Traditional water quality assessment methods rely on in-situ sampling and laboratory analysis, which, while accurate, are often time-consuming, resource-intensive, and spatially limited. HydroVision introduces a scalable, non-invasive alternative by leveraging CNN and transformer-based models to extract meaningful insights from publicly available imagery.

One of the primary challenges in using RGB images for water quality estimation is the absence of spectral information beyond the visible range, which limits direct measurement of parameters such as chlorophyll concentration and turbidity. However, our study demonstrates that deep learning models can learn meaningful correlations between visible water features—such as color intensity, surface texture, and reflectance patterns—and optically active parameters when trained on well-processed datasets.

The results indicated that DenseNet121 outperformed other CNNs, achieving the highest validation R² values for CDOM (0.898), Chlorophyll-$\alpha$ (0.678), and Chlorophylls (0.788), highlighting its ability to capture fine-grained spatial variations. Additionally, the Vision Transformer (ViT) showed potential for Suspended Sediments, achieving the highest Pearson correlation coefficient (0.242) among all models, though with limited agreement and higher bias, likely due to data imbalance and ViT's sensitivity to small datasets.

While RGB-based AI models show strong potential for remote water quality assessment, their performance remains dependent on dataset quality, environmental conditions, and parameter-specific sensitivities. Future research should focus on expanding labeled datasets and exploring advanced architectures to further improve generalizability across diverse aquatic environments.

\subsection{Broader Implications and Limitations}
The HydroVision model has the potential to benefit a diverse range of stakeholders by providing a scalable, cost-effective approach to water quality monitoring. Environmental agencies and water management authorities, such as the U.S. Environmental Protection Agency (EPA), the Virginia Department of Environmental Quality (DEQ), and the Association of Clean Water Administrators (ACWA), can leverage HydroVision to improve real-time water quality assessments, improve resource management, and mitigate contamination risks to prevent emergencies such as boil water alerts \cite{virginia_deq,dcwater_boil_advisory}. Regulatory bodies and industrial compliance authorities overseeing wastewater discharge from manufacturing plants and chemical facilities can utilize AI-powered solutions like HydroVision to ensure compliance with environmental regulations, detect pollution events in real-time, and respond effectively to force majeure events \cite{nytimes2025planecrash}. Additionally, the National Park Service (NPS) and conservation organizations can integrate this model into ecosystem monitoring efforts, ensuring the sustainability of protected water bodies and enhancing visitor safety. Outdoor enthusiasts, such as hikers and adventurers, can also benefit from real-time contamination alerts, enabling informed decisions about water use, particularly in regions prone to natural disasters like landslides and tsunamis \cite{nps_water_quality,cleanwater_appalachian_trail,10.1145/3708359.3712107}.

Despite its potential, this study has several limitations that should be considered for future research. First, the model relies exclusively on RGB images, which limits its ability to detect non-visible water quality indicators such as chemical contaminants and heavy metals. Integrating hyperspectral or multispectral imaging could improve prediction accuracy for a broader range of parameters. Second, RGB-based predictions are inherently sensitive to environmental lighting conditions—such as low illumination, nighttime, or weather-obstructed scenes (e.g., fog, rain, or glare)—which may degrade model performance and introduce uncertainty. Incorporating data normalization techniques, robust augmentation strategies, or low-light image enhancement methods could help mitigate these challenges. Third, the effectiveness of the U-Net segmentation model varies depending on environmental conditions, with occasional misclassification of sky regions as water, which may introduce errors in downstream predictions. While filtering techniques mitigate this issue, further refinements in segmentation models—such as incorporating additional training data or hybrid deep-learning approaches—could enhance robustness. Finally, while CNN models demonstrate strong predictive capabilities, their interpretability remains a challenge, particularly for regulatory adoption. Future work should focus on improving the generalization and interpretability of AI-driven water quality models. This includes enhancing data augmentation strategies, increasing feature diversity, and exploring advanced architectural tuning—such as the integration of spectral (e.g., NIR, hyperspectral) or temporal data—to better capture visually ambiguous parameters. Additionally, incorporating explainability techniques will be essential to ensure that model predictions can be reliably interpreted and trusted by policymakers, environmental agencies, and other stakeholders.

\section*{Acknowledgment}
This research would not have been possible without inputs from our colleagues at the A3 Lab at Virginia Tech (\url{https://ai.bse.vt.edu/}), and for fostering an environment of innovation and collaboration. We also acknowledge the Commonwealth Cyber Initiative (CCI) for the continuous support. We also acknowledge the United States Geological Survey (USGS) for providing open access to the HIVIS and NWIS datasets, which were essential to this work.

\section*{Data Availability Statement}
The HydroVision dataset, including segmented images and associated water quality parameters, 
is available at the \href{https://ai.bse.vt.edu/}{\uline{\textcolor{blue}{A3 lab}}}'s open access repository. Additional data or processing scripts are available upon request from the corresponding author.

\sloppy
\bibliographystyle{IEEEtran}

\newpage
\section{Authors Biographies}

\vspace{11pt}

\begin{IEEEbiography}[{\includegraphics[width=1in,height=1.25in,clip,keepaspectratio]{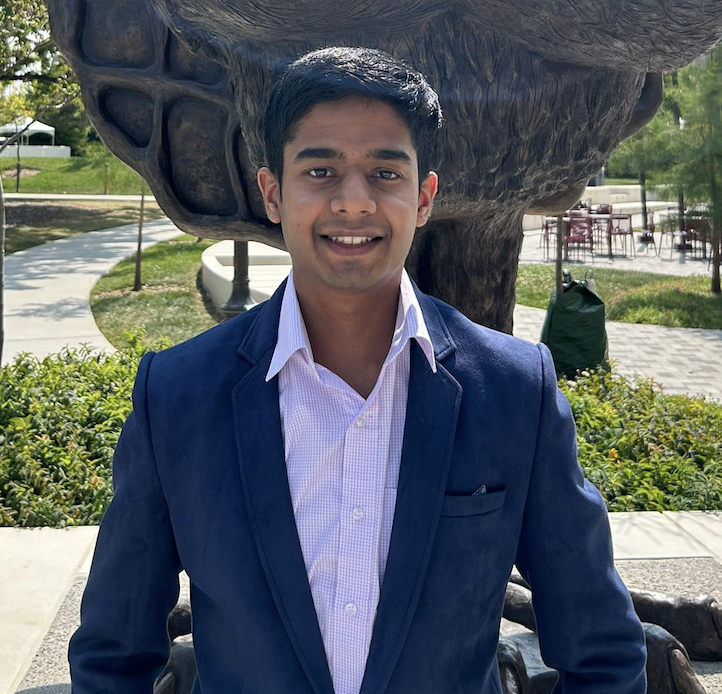}}]{Shubham Laxmikant Deshmukh}
is a graduate student in Computer Science at Virginia Tech with a specialization in Artificial Intelligence. His research interests include computer vision, deep learning, and environmental monitoring, with a focus on developing scalable, AI-driven solutions for water quality assessment. He currently serves as a Graduate Research Assistant at the A3 Lab within the Commonwealth Cyber Initiative (CCI), where he has designed and deployed deep learning pipelines using U-Net and Vision Transformer models for surface water parameter prediction, significantly improving prediction accuracy and computational efficiency.
\end{IEEEbiography}

\vspace{11pt}

\begin{IEEEbiography}[{\includegraphics[width=1in,height=1.25in,clip,keepaspectratio]{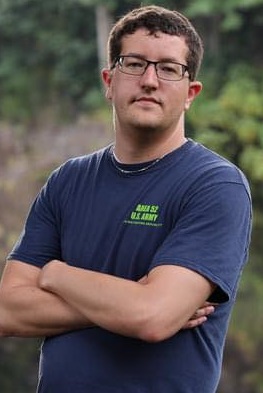}}]{Matthew Wilchek}
is a Ph.D. candidate in Computer Science at Virginia Tech and a Data Scientist at the U.S. Army DEVCOM C5ISR Center. His research focuses on human-centered artificial intelligence, computer vision, augmented reality, and AI assurance, with applications to national security and cyber-physical systems. He has published in venues such as ACM Transactions on Interactive Intelligent Systems, ACM Intelligent User Interfaces, and Engineering Applications of Artificial Intelligence. He has experience leading interdisciplinary research studies involving human-in-the-loop and crowd-in-the-loop systems in both simulated and field environments. His work bridges operational defense challenges and experimental AI systems research.
\end{IEEEbiography}

\vspace{11pt}

\begin{IEEEbiography}[{\includegraphics[width=1in,height=1.25in,clip,keepaspectratio]{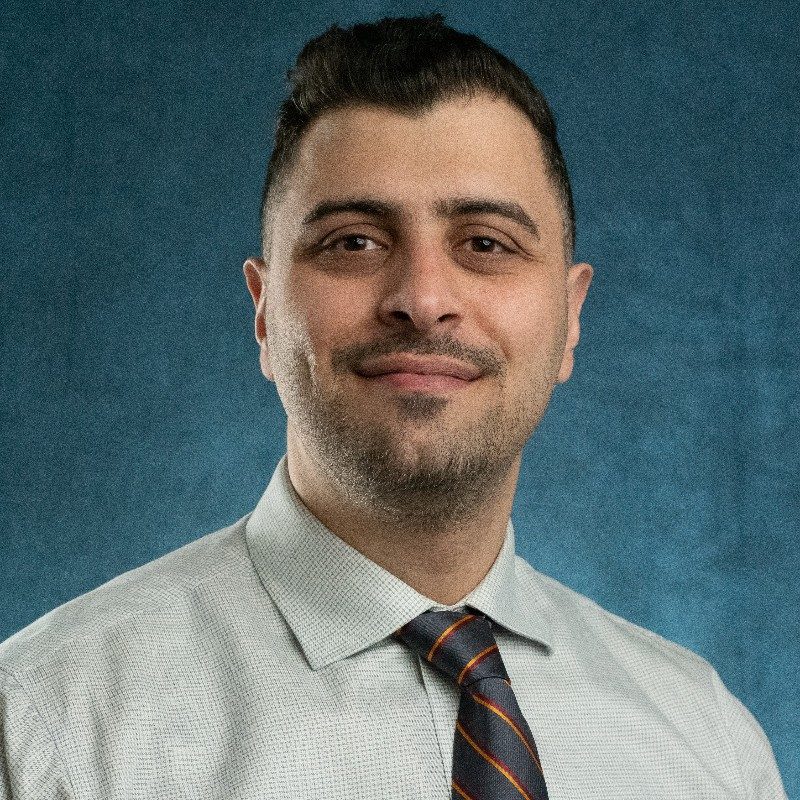}}]{Feras A. Batarseh}
is an associate professor with the Department of Biological Systems Engineering and Commonwealth Cyber Initiative (CCI) at Virginia Tech (VT); he is the Director of A3 (AI Assurance and Applications) Lab. His research spans the areas of Trustworthy AI, Intelligent Water Systems, Cyberbiosecurity, and Data-Driven Public Policy.
Dr. Batarseh published 100+ conference and journal papers at prestigious venues, additionally, he has co-authored three books: "Federal Data Science", "Data Democracy", and "AI Assurance", all published by Elsevier’s Academic Press. Dr. Batarseh is a senior member of the Institute of Electrical and Electronics Engineers (IEEE), and a member of the Agricultural and Applied Economical Association (AAEA), and the Association for the Advancement of Artificial Intelligence (AAAI). He has taught AI courses at multiple universities including George Mason University (GMU), University of Maryland - Baltimore County (UMBC), Georgetown University, and George Washington University (GWU). Dr. Batarseh obtained his Ph.D. and M.Sc. in Computer Engineering from the University of Central Florida (UCF) (2007, 2011), a Juris Masters of Law from GMU (2022), and a Graduate Certificate in Project Leadership from Cornell University (2016). 

\end{IEEEbiography}

\vfill

%%%%%%%%%%%%%%%%%%%%%%%%%%%%%%%%%%%%%%%%%%%%%%%%%%%%%%

\end{document}